\documentclass[sigconf]{acmart}

\usepackage{booktabs} % For formal tables

\usepackage{latexsym}
\usepackage{graphicx}
\usepackage{CJKutf8}
\usepackage{epsfig}
\usepackage{algorithm, algpseudocode}
\usepackage{algpseudocode}
\usepackage{booktabs}
\usepackage{multirow}

\usepackage{makecell}
\usepackage{color}
\usepackage{subfigure}
\usepackage{url}
\usepackage{mathrsfs}
\usepackage{amsmath}

\newcommand{\secref}[1]{Section \ref{#1}}
\newcommand{\figref}[1]{Figure \ref{#1}}

\newcommand{\tabref}[1]{Table \ref{#1}}

\newcommand{\argmax}{\operatornamewithlimits{argmax}}

\begin{document}

\copyrightyear{2018} 
\acmYear{2018} 
\setcopyright{acmlicensed}
\acmConference[CIKM '18]{The 27th ACM International Conference on Information and Knowledge Management}{October 22--26, 2018}{Torino, Italy}
\acmBooktitle{The 27th ACM International Conference on Information and Knowledge Management (CIKM '18), October 22--26, 2018, Torino, Italy}
\acmPrice{15.00}
\acmDOI{10.1145/3269206.3269326}
\acmISBN{978-1-4503-6014-2/18/10}

\title{Query Tracking for E-commerce Conversational Search: A Machine Comprehension Perspective}

%\subtitle{Extended Abstract}

\author{
	Yunlun Yang, 
	Yu Gong, 
	Xi Chen
}
\affiliation{
  \institution{Search Algorithm Team, Alibaba Group}}
\email{{yunlun.yyl, gongyu.gy}@alibaba-inc.com, gongda.cx@taobao.com}

\renewcommand{\shorttitle}{Query Tracking for E-commerce Conversational Search: \\ A Machine Comprehension Perspective}

\begin{abstract}
With the development of dialog techniques,
conversational search has attracted more and more attention as it enables users to interact with the search engine in a natural and efficient manner.
However, comparing with the natural language understanding in traditional task-oriented dialog which focuses on slot filling and tracking,
the query understanding in E-commerce conversational search is quite different and more challenging due to more diverse user expressions and complex intentions.
In this work, we define the real-world problem of query tracking in E-commerce conversational search, in which the goal is to update the internal query after each round of interaction.
We also propose a self attention based neural network to handle the task in a machine comprehension perspective.
Further more we build a novel E-commerce query tracking dataset from an operational E-commerce Search Engine,
and experimental results on this dataset suggest that our proposed model outperforms several baseline methods by a substantial gain for Exact Match accuracy and F1 score, showing the potential of machine comprehension like model for this task.
\end{abstract}

\begin{CCSXML}
<ccs2012>
<concept>
<concept_id>10002951.10003317.10003325</concept_id>
<concept_desc>Information systems~Information retrieval query processing</concept_desc>
<concept_significance>500</concept_significance>
</concept>
<concept>
<concept_id>10002951.10003317.10003331</concept_id>
<concept_desc>Information systems~Users and interactive retrieval</concept_desc>
<concept_significance>500</concept_significance>
</concept>
</ccs2012>
\end{CCSXML}

\ccsdesc[500]{Information systems~Information retrieval query processing}
\ccsdesc[500]{Information systems~Users and interactive retrieval}

\keywords{query tracking, machine comprehension, neural network, conversational search}

\maketitle

\renewcommand{\thefootnote}{\fnsymbol{footnote}}
%\footnotetext[1]{Equally contribution.}
%\footnotetext[2]{Corresponding author.}
\renewcommand{\thefootnote}{\arabic{footnote}}

\begin{CJK}{UTF8}{gbsn}
	
	\section{Introduction}
\label{sec:intro}

Searching is often the first step for a user when he want to seek specific products on an E-commerce website. Traditional search interface restricts the user to search statelessly, and the user has to manually revise the query based on the results in last search. 
However, we find that consecutive queries issued in a short period are typically related. A basic example is asking ``dress'' followed by ``red dress''.
Therefore making good use of the context queries in previous search may bring more efficient interaction.

With the renaissance of neural network and the development of conversation techniques, interaction between humans and machines in a conversational manner has become popular, e.g. smart assistants, chatbots for chit-chat and task-oriented dialog \cite{li2017end,li2016persona,zhu2017next,wang2016view}.
By adopting the conversation ideas in search systems, the so-called ``conversational search'' \cite{radlinski2017theoretical} takes the user's latest queries into consideration and enables a user to express his new search intention in a naturally chatting way,
which makes the entire search process more fluent.

Conversational search in E-commerce and task-oriented dialog share the same goal of understanding users' demands and helping complete corresponding tasks \cite{yan2017building}.
For example, both the dialog system for flight reservation and E-commerce conversational search will chat with the user
and acquire his preferences, such as flight destination and item property respectively,
finally provide the user expected flight tickets or products.
The performance of such systems significantly relies on input understanding module.

In task-oriented dialog systems, natural language understanding mainly composes of two components: slot filling and dialog state tracking.
Slot filling aims at extracting slot values and corresponding slots in input utterances. Dialog state tracker monitors the user's current intention by maintaining a compact representation, which is usually called ``dialog state'', and takes the results of slot filling into account to update the dialog states.
The dialog states are further fed into the dialog manager module to decide the subsequent interaction.

Nevertheless, the query understanding of conversational search in E-commerce is of great difference with the natural language understanding in task-oriented dialog system.
First, in a dialog system, the complexity of user utterance is often proportional to the diversity of what the system can offer to users. Since E-commerce search systems aim to help consumers find their desired items from a large variety of products, consumers' descriptions for products are supposed to be particularly diverse.
If slot filling is applied, large amount of semantic slots should be defined to cover all product properties, e.g. brand, color, size, style, skirt length and so on, 
and it is an expensive manual work.
Second, unlike task-oriented dialog, a user may try different colors, brands and other keywords to approach satisfied item when interacting with search systems,
and this raises high requirements to the state tracking module.
For instance, if a user first searches for ``Adidas shoes'', and then types in ``Nike'', his current intention should be ``Nike shoes'' since both ``Adidas'' and ``Nike'' are tagged as ``brand'' by slot filling , but if first input is ``fairy dress'' and then ``cute'', the user probably wants to search for ``fairy and cute dress'', although both ``fairy'' and ``cute'' are words that describe the style of clothes. Hence, simple slot based state tracking may not be able to handle complex situations in E-commerce conversational search.
Last but not the least, it is important to summarize current user intention in a query and feed it to search engine in conversational search \cite{ren2018conversational,kumar2017incomplete}. We call the query that represents current user intention ``internal query''. Comparing with task-oriented dialog, the internal query plays the role of ``user state'', and is updated after each round of interaction. It is noteworthy that queries in E-commerce search commonly only contain several unordered keywords that describe some key properties of user's needed products so as the internal query. An example of user input queries and corresponding internal queries are shown in \tabref{table:example}.

\begin{table}[htbp]
	\centering
	\small
	\caption{An example of user input queries and corresponding internal queries.}
	\begin{tabular}{|c|l|l|}
		\hline
		\#Turn & User Input Query & Internal Query \\
		\hline
		1 & sport shoes & sport shoes \\
		\hline
		2 & Adidas & Adidas sport shoes \\
		\hline
		3 & Nike black & black Nike sport shoes \\
		\hline
		4 & ventilated & ventilated black Nike sport shoes \\
		\hline
	\end{tabular}
	\label{table:example}
	\vspace{-10pt}
\end{table}

Therefore, to attack the challenges mentioned above and improve the tracking of the internal query in E-commerce conversational search, We formulate the problem as a context-aware query tracking task. The goal is to consider a user's query history in a conversational way and output a query indicating the user's current intention.
To handle this task, we involve a machine comprehension perspective, and propose a neural network based query tracking model. In our model, we update the internal query without the help of slot, and turn the problem to a word-level binary classification. 

\iffalse
The main contributions of our work are:

\begin{itemize}
%\itemsep0em
\item We define a real-world task of context-aware query tracking for the E-commerce conversational search system (\secref{sec:problem}),
	and we propose a novel neural network model based on machine comprehension perspective (\secref{sec:model}).
\item We develop a Chinese E-commerce conversational search dataset for query tracking from online search engine query logs.
\item Multiple experiments suggest that our proposed model outperforms several baseline methods showing the potential of machine comprehension like model for this task.
\end{itemize}
\fi

The main contributions of our work are: 1) We define a real-world task of context-aware query tracking for the E-commerce conversational search system (\secref{sec:problem}), and we propose a novel neural network model based on machine comprehension perspective (\secref{sec:model}).
2) We develop a Chinese E-commerce conversational search dataset for query tracking from online search engine query logs.
3) Multiple experiments suggest that our proposed model outperforms several baseline methods showing the potential of machine comprehension like model for this task.

	\section{Problem}
\label{sec:problem}
In E-commerce conversational search, at each turn of interaction, a user types in a query, and the goal of query tracking is to consider the user's query history in a conversational way and provide a query indicating the user's current intention. The output query is also called "internal query", and will be further fed to the search engine to get search results. Assuming that the update of internal query satisfies Markov property, i.e. the user changes his intention only based on the last search, we can get the new internal query by considering only the last internal query and current input query.

Formally, given the last internal query $\textbf{q}_1 = \{w^1_t\}_{t=1}^m$, and the current input query $\textbf{q}_2 = \{w^2_t\}_{t=1}^n$, the goal of our problem is to generate the new internal query $\hat{\textbf{q}}_3 = \{w^3_t\}_{t=1}^k$ such that 
\begin{equation*}
	\hat{\textbf{q}}_3 = \mathop{\arg\max}_{\textbf{q}_3}P(\textbf{q}_3|\textbf{q}_1, \textbf{q}_2)
\end{equation*}
\noindent Which is sequence generation task. It is noteworthy that the words in $\textbf{q}_2$ must appear in $\hat{\textbf{q}}_3$. And the words in query are assumed unordered since the word order of queries in E-commerce search hardly affects the results of search engine. Based on these observations, we can solve the task by deciding whether each word in $\textbf{q}_1$ should be reserved and taking the union set of remaining words of $\textbf{q}_1$ and $\textbf{q}_2$ as $\hat{\textbf{q}}_3$. Formally, the goal turns to finding a sequence of labels $\hat{\textbf{y}} = (y_1, y_2, \ldots, y_m)$, one for each word in the last internal query, such that
\begin{equation*}
	\hat{\textbf{y}} = \mathop{\arg\max}_{\textbf{y}}P(\textbf{y}|\textbf{q}_1, \textbf{q}_2)
\end{equation*}
\noindent where $y_i \in \{0, 1\}$ and indicates whether the word $w^1_i$ should be reserved or discarded in the new internal query.

	\section{Approach}
\label{sec:model}

In this section, we first discuss the connections between query tracking and machine reading as well as their main differences.
And then motivated by ideas from machine reading, we propose a neural network based query tracking model.

\subsection{Machine Comprehension V.S. Query Tracking}
Machine reading has become a research hotspot recently. In this task, given a document, the goal is to answer a question related to the document \cite{wang2016machine}. Therefore, each data sample in machine reading is a triplet, $(Doc, Que, Ans)$. The framework of mainstream machine comprehension model typically consists of three phases: Encoding, Matching and Predicting. High-level representations of all words in documents and questions are produced in Encoding phase, and the question features are fused into document features by attention-based word pair matching in Matching phase. At last, the model will directly predict answer, which is typically a word or span of the document.

The goal of query tracking is similar to that of machine reading: given the current input query, find the snippets of the last internal query that should be reserved. Let $\textbf{q}_1'$ denote the reserved words from $\textbf{q}_1$, then a data sample in query track is also a triplet, $(\textbf{q}_1, \textbf{q}_2, \textbf{q}_1')$, in which $\textbf{q}_1'$ can be further translated into one-hot labels $\textbf{y}$.

The differences are obvious as well. First, answers in machine comprehension must be a word or a sequence of consecutive tokens from the documents, whereas the reserved words are not guaranteed sequential. As a result, directly predicting the word or answer span is no more suitable for query tracking. Second, as stated above, words in queries are assumed unordered, which means a query is actually a bag of words and representing queries with sequential models like LSTM-RNN may cause data inefficiency when training. Third, query tracking in a search session is a recurrent process and previous query tracking error may propagate through the subsequent interactions, hence the performance of query tracking should be more reliable than machine comprehension.

Overall, we follow the basic framework of machine comprehension and build our model for query tracking task.

\subsection{Proposed Model}
\begin{figure}[th]
\centering  
\includegraphics[angle=0,width=1.0\columnwidth]{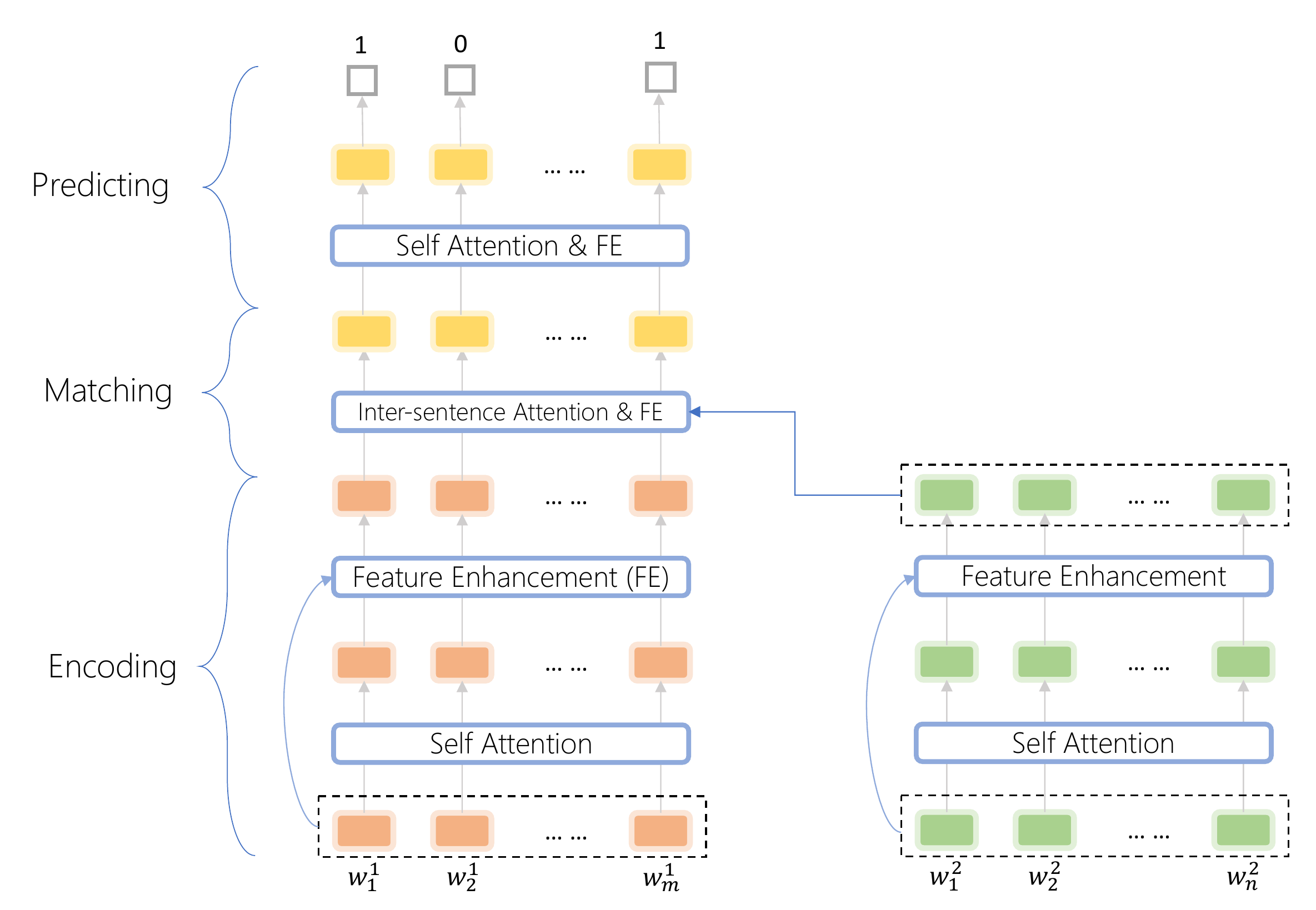}  
\caption{The framework of our proposed model.}
\label{fig:framework}
\vspace{-10pt}
\end{figure}

Motivated by the connections and differences between machine reading and query tracking, we propose a neural network based model to handle this task, which is composed of the same three parts: Encoding, Matching, Predicting, as shown in \figref{fig:framework}.
\vspace{-2mm}
\subsubsection{Encoding}
Consider the last internal query $\textbf{q}_1 = \{w_t^1\}_{t=1}^m$ and the current input query $\textbf{q}_2 = \{w_t^2\}_{t=1}^n$. First the words are converted to their respective word embeddings, $\{e_t^1\}_{t=1}^m$ and $\{e_t^2\}_{t=1}^n$, where $e_t^1, e_t^2 \in \mathbb{R}^{d_w}$. Typically, these word embedding will be further fed to a bi-directional recurrent neural network to get higher-level representations in machine reading. However, considering the assumption that a query is a bag of unordered words, applying the RNN model which takes sequential information into account may hurt the training efficiency. In query tracking, we propose to leverage self attention based encoder as a more suitable substitute. For a query whose embedding matrix is $E \in \mathbb{R}^{l \times d_w}$, its encoding process is formulated as:
\begin{equation*}
SelfAttention(E) = softmax(\frac{EE^\mathrm{T}}{\sqrt{d_w}})E
\end{equation*}

\noindent Where $softmax$ is row-wise and we use scaled dot-product as attention scoring function. More specifically, we exploit the multi-head mechanism \cite{vaswani2017attention} as well:
\begin{align*}
SelfAttentionMH(E) & = Concat(head_1, \ldots, head_h)W_{mh} \\
\text{where } head_i & = SelfAttention(EW_i)
\end{align*}

\noindent Where $h$ is the number of heads, $W_i \in \mathbb{R}^{d_w \times d_h}$, $W_{mh} \in \mathbb{R}^{hd_h \times d_w}$ and $d_h$ is the word dimension in each head.

We also employ a feature enhancement technique \cite{mou2016natural} widely used in sentence pair modeling. Supposing that $H = SelfAttentionMH(E)$, we concatenate the features and feed the result to a fully-connected layer:
\begin{equation*}
M = f([E, H, E - H, E \circ H]W_{fe})
\end{equation*}

\noindent Where $\circ$ denotes element-wise product, $f(\cdot)$ denotes activation function, and $W_{fe} \in \mathbb{R}^{4d_w \times d_w}$. The element-wise difference and product can be regarded as a strong prior for $W_{fe}$ and intuitively help capture the relation between words. Comparing with simple addition (residual connection in Transformer) and concatenation, the feature enhancement brings better performance in experiments. After Encoding, the representations of $\textbf{q}_1$ and $\textbf{q}_2$ are denoted as $M_1$ and $M_2$ respectively.

\subsubsection{Matching}
Similar to fusing passage and question representations in machine reading, we aim to incorporate the information of current input query into the last internal query in Matching phase. In query tracking, for determining whether words in internal query should be discarded, it is important to check for contradictions\footnote{"Contradiction" here usually means belonging to the same property, like "Adidas" and "Nike", "red" and "black".} of individual word pairs between queries. To realize the idea, we utilize word-by-word attention \cite{bahdanau2014neural} to make information flow between queries. Formally, 
\begin{gather*}
Attention(M_1, M_2) = softmax(\frac{M_1M_2^\mathrm{T}}{\sqrt{d_w}})M1 \\
Y = Concat(head_1, \ldots, head_h)W_{mh} \\
\text{where } head_i = Attention(M_1W^1_i, M_2W^2_i)
\end{gather*}

Where $W^1_i, W^2_i \in \mathbb{R}^{d_w \times d_h}$. The feature enhancement is applied as well:
\begin{equation*}
M_3 = f([M_1, Y, M_1 - Y, M_1 \circ Y]W_{fe})
\end{equation*}

Since self attention is a special case of inter-sentence attention, their implementations are similar.In Matching phase, the words in $\textbf{q}_1$ are expected to find and absorb their counterpart information in $\textbf{q}_2$ by word-by-word attention.

\subsubsection{Predicting}
In machine comprehension, answers are extracted generally by selecting the word or span with high confidence in documents. Due to the difference that reserved word may consist of inconsecutive snippets, we employ a straightforward strategy and directly predict the labels of words, i.e. binary classification for each word. Before predicting, we apply self attention based encoding again to $M_3$ to get $M_4$. Experimental results show that the second encoding operation helps words in the same phrase (e.g. brand "vero moda") behave consistently in classification. Based on $M_4$, the binary classification are formulated as:
\begin{equation*}
\textbf{y} = \argmax softmax(M_4W_{bc})
\end{equation*}

\noindent Where $W_{bc} \in \mathbb{R}^{d_w \times 2}$, and $\textbf{y}$ is the predicted label. We train the network by minimizing the sum of cross entropy between ground truth label and predicted distribution.
	
	\section{Experiments}
\label{sec:exp}
%In this section, we first introduce how we build the query tracking dataset. And the second part is about the experimental settings and implementation details of proposed approach. At last, we compare our method with several baselines and report their performances.

\subsection{Dataset}
To collect reliable query tracking data, we extract user input query from the query log of an online Chinese E-commerce search system. Since the online search system is not conversational and users interact with it in a single-turn way, we need to mine the change of user intentions from user input session. 

First, query pairs which are typed in consecutively by the same user within $30$ minutes are extracted, and those of which frequencies are less than $5$ are filtered out. We regard the two queries in a pair are respectively the internal queries of two consecutive interaction round in conversational search, i.e. $\textbf{q}_1$ and $\hat{\textbf{q}}_3$ in our definition. And Chinese word segmentation is applied to these queries. Then, we consider the difference set of the words in two queries of a pair as $\textbf{q}_2$. For example, supposing the query pair (translated into English) is ("Adidas shoes", "Nike shoes"), the generated data triplet $(\textbf{q}_1, \textbf{q}_2, \hat{\textbf{q}}_3)$ is ("Adidas shoes", "Nike", "Nike shoes"). At last, the triplets of which $\textbf{q}_2$ is empty or meaningless are removed, and we split the rest into three parts, get $4,865$/$271$/$275$ thousand samples as train/validation/test set.

\subsection{Implementation Details}
Word embeddings used in our model are initialized by pretraining on a large shopping related corpus and fine-tuned during training. The head number and dimension of multi-head attention are $5$ and $40$. 
All queries are padded to a maximum sequence length of $20$.
We use Adam as the optimization algorithm.
The initial learning rate is $0.001$ and further decayed exponentially.
For regularization, we apply dropout to the word embedding and the output of attention layer with a ratio of $0.1$, and conduct early stopping on validation set.

\subsection{Main Results}
We evaluate our proposed model and several baseline methods, including:

\begin{itemize}
\item \textbf{Slot baseline} We use a well-maintained product attribute knowledge base as the slot definition and develop a dynamic programming algorithm to match the slots in queries. In tracking, new value will replace the old one for the same slot. All slot values are concatenated as $\hat{\textbf{q}}_3$.
\item \textbf{LSTM baseline 1} Model both queries with LSTM and feed the final state of $\textbf{q}_1$ as the initial state of $\textbf{q}_2$.
\item \textbf{LSTM baseline 2} Use bidirectional LSTM as encoder and the rest parts is the same as our proposed model.
\end{itemize}

We report the query-level Exactly Matched accuracy and word-level F1 score as evaluation metric.
\vspace{-2mm}
\begin{table}[ht]
	\centering
	\small
	\caption{Performances of baseline methods and proposed model on query tracking dataset.}
	\begin{tabular}{c|cc}
		\toprule
		Methods & EM & F1 \\
		\midrule
		Slot baseline & 66.7 & 72.1 \\
		LSTM baseline 1 & 85.1 & 90.5 \\
		LSTM baseline 2 & 86.8 & 91.5 \\
		\midrule
		Our proposed model & \textbf{86.9} & \textbf{91.6} \\
		\bottomrule
	\end{tabular}
	\label{table:overall_results}
	\vspace{-10pt}
\end{table}
%\vspace{-3mm}
\tabref{table:overall_results} is the results of baseline methods and proposed model on query tracking dataset. The unsupervised slot based method is limited by the coverage of product knowledge base and the effectivity of slot filling algorithm, and yields a much worse result than the supervised methods. This suggests that the E-commerce conversational search is a complex scenario and requires abundant knowledge and elaborate tracking rules to handle it in an unsupervised way. The other two strong baselines are both based on LSTM RNN encoding. LSTM baseline 1 outperforms Slot baseline by a large margin. LSTM baseline 2 achieves a comparable result with our proposed model. However, taking model efficiency into account, our proposed model is $2.1\times$ faster than LSTM baseline 2 in training speed and $1.6\times$ in test speed. While yielding close results, our self attention based model is more efficient.  
We also modify our model by adding a bidirectional LSTM layer before self attention encoding to increase the capacity, but no further improvement is observed.

\subsection{Ablation Test}
\vspace{-3mm}
\begin{table}[ht]
	\centering
	\small
	\caption{Ablation test results of proposed model on query tracking dataset.}
	\begin{tabular}{l|cc}
		\toprule
		Methods & EM & F1 \\
		\midrule
		Full model & 86.9 & 91.5 \\
		\midrule
		- pretrained embedding (randomly initialized instead) & 86.0 & 90.9 \\
		- self attention based encoding & 85.8 & 90.8 \\
		- multi-head mechanism & 86.4 & 91.1\\
		- feature enhancement (concatenation instead) &  86.5 & 91.2 \\
		- feature enhancement (addition instead) & 86.1 & 91.0 \\
		\bottomrule
	\end{tabular}
	\label{table:ablation_test}
	\vspace{-10pt}
\end{table}
%\vspace{-3mm}
The results of ablation test are shown in \tabref{table:ablation_test}. Self attention based encoding is crucial to the performance of our model, and only using word embedding will hurt the EM accuracy by $1.1$. With pretrained word embedding, the EM accuracy improves from $86.0$ to $86.9$, suggesting that well-initialized parameters are helpful to the final results. We have also tried other feature combination strategies and find element-wise product and difference are more important features. And the experimental results support that multi-head is a useful mechanism in attention.

	\section{Conclusion}
\label{sec:conclusion}

In this paper, We define a real-world task of context-aware query tracking for the E-commerce conversational search system, and propose a novel attention based neural network model on machine comprehension perspective. For evaluation, we develop a Chinese E-commerce conversational search dataset for query tracking from online search engine query logs. And experiments suggest that our proposed model outperforms several baseline methods showing the potential of machine comprehension like model for this task and the efficiency of attention based model.
%For future work, we will consider the output results of slot filling as additional features and integrate them into neural network. And improving the network architecture is a possible direction as well.
	
\end{CJK}

\bibliographystyle{ACM-Reference-Format}
\bibliography{ref}

\end{document}